\documentclass[10pt,twocolumn,letterpaper]{article}
\usepackage[table]{xcolor}
\usepackage{cvpr}
\usepackage{times}
\usepackage{epsfig}
\usepackage{graphicx}
\usepackage{adjustbox}
\usepackage{amsmath}
\usepackage{tabularx}
\usepackage{amssymb}
\usepackage{stmaryrd}
\usepackage{mathtools}
\usepackage{float}
\usepackage{tabu}
\usepackage{multirow}
\usepackage{hhline}
\usepackage{url}
\usepackage{flushend}
\usepackage[mathcal]{euscript}

\newcommand{\patrev}[1]{\textcolor{black}{{#1}}}

\renewcommand{\paragraph}[1]{\medskip\noindent\textbf{#1}\enskip}

\newcommand{\coco}{{\tt MS-COCO}}
\newcommand{\flickr}{{\tt Flickr-30K}}
\newcommand{\genome}{{\tt Visual Genome}}
\newcommand{\si}[2]{\langle #1,#2 \rangle}

\usepackage[pagebackref=true,breaklinks=true,letterpaper=true,colorlinks,bookmarks=false]{hyperref}

\cvprfinalcopy 

\def\cvprPaperID{3272} 

\begin{document}
\newcommand{\mbf}[1]{\ensuremath{\mathbf{#1}}}
\newcommand{\mbs}[1]{\ensuremath{\boldsymbol{#1}}}
\newcommand{\mcl}[1]{\ensuremath{\mathcal{#1}}}
\newcommand{\mrm}[1]{\ensuremath{\mathrm{#1}}}
\newcommand{\mbb}[1]{\ensuremath{\mathbb{#1}}}
\newcommand{\msf}[1]{\ensuremath{\mathsf{#1}}}

\newcommand{\ve}[1]{\ensuremath{\mathbf{#1}}} 
\newcommand{\m}[1]{\ensuremath{\mathsf{#1}}} 
\newcommand{\s}[1]{\ensuremath{\mathcal{#1}}} 

\newcommand{\trsp}[1]{\ensuremath{#1^{\top}}}
\newcommand{\pinv}[1]{\ensuremath{#1^{\dagger}}}
\newcommand{\bmat}[4]{\ensuremath{\begin{bmatrix}#1&#2\\#3&#4\end{bmatrix}}}
\def\trace{\ensuremath{\mathrm{trace}}}
\def\deter{\ensuremath{\mathrm{det}}}
\def\diag{\ensuremath{\mathrm{diag}}}
\def\rank{\ensuremath{\mathrm{rank}}}
\def\Id{\m{Id}} 
\def\mA{\m{A}}
\def\mB{\m{B}}
\def\mC{\m{C}}
\def\mD{\m{D}}
\def\mE{\m{E}}
\def\mF{\m{F}}
\def\mG{\m{G}}
\def\mH{\m{H}}
\def\mK{\m{K}}
\def\mL{\m{L}}
\def\mN{\m{M}}
\def\mP{\m{P}}
\def\mW{\m{W}}
\def\mX{\m{X}}
\def\mY{\m{Y}}
\def\mZ{\m{Z}}

\newcommand{\bvec}[2]{\ensuremath{\begin{bmatrix}#1\\#2\end{bmatrix}}}
\def\One{\mbs{1}} 
\def\va{\ve{a}}
\def\vb{\ve{b}}
\def\vc{\ve{c}}
\def\vd{\ve{d}}
\def\vf{\ve{f}}
\def\vg{\ve{g}}
\def\vh{\ve{h}}
\def\vi{\ve{i}}
\def\vt{\ve{t}}
\def\bx{\ve{x}}
\def\by{\ve{y}}
\def\bz{\ve{z}}
\def\bv{\ve{v}}

\def\cL{\mcl{L}}

\def\ie{\emph{i.e.}}
\def\eg{\emph{e.g.}}
\def\iid{\emph{i.i.d.}}
\def\wrt{w.r.t.}
\def\mwrt{\mrm{w.r.t.}}
\def\msbt{\mrm{sb.t.}}

\def\sqt{^{\frac{1}{2}}} 
\def\msqt{^{-\frac{1}{2}}} 
\def\R{\mbb R}
\def\vtheta{\mbs{\theta}}

\title{Finding beans in burgers:\\Deep semantic-visual embedding with localization}

\author{Martin Engilberge\textsuperscript{1,2}, Louis Chevallier\textsuperscript{2}, Patrick P{\'e}rez\textsuperscript{2}, Matthieu Cord\textsuperscript{1}\\[1ex]
\textsuperscript{1}Sorbonne universit\'e, Paris, France
\textsuperscript{2}Technicolor, Cesson S\'evign\'e, France\\[1ex]
{\tt\small \{martin.engilberge, matthieu.cord\}@lip6.fr}  \\
{\tt\small patrick.perez@valeo.com  louis.chevallier@technicolor.com}
}

\maketitle

\begin{abstract}

Several works have proposed to learn a two-path neural network that maps images and texts, respectively, to a same shared Euclidean space where geometry captures useful semantic relationships. Such a multi-modal embedding can be trained and used for various tasks, notably image captioning. In the present work, we introduce a new architecture of this type, with a visual path that leverages recent space-aware pooling mechanisms. Combined with a textual path which is jointly trained from scratch, our semantic-visual embedding offers a versatile model. Once trained under the supervision of captioned images, it yields new state-of-the-art performance on cross-modal retrieval. It also allows the localization of new concepts from the embedding space into any input image, delivering state-of-the-art result on the visual grounding of phrases.   

\end{abstract}

\section{Introduction}\label{sec:intro}
Text and image understanding is progressing fast thanks to the ability of artificial neural nets to learn, with or without supervision, powerful distributed representations of input data. At runtime, such nets \textit{embed} data into high-dimensional feature spaces where semantic relationships are geometrically captured and can be exploited to accomplish various tasks. Off-the-shelf already trained nets are now routinely used to extract versatile deep features from images 
which can be used for recognition or editing tasks, or to turn words and sentences into vectorial representations that can be mathematically analysed and manipulated.

\begin{figure}[htb]
 \begin{center}
  \includegraphics[width=1.0\linewidth]{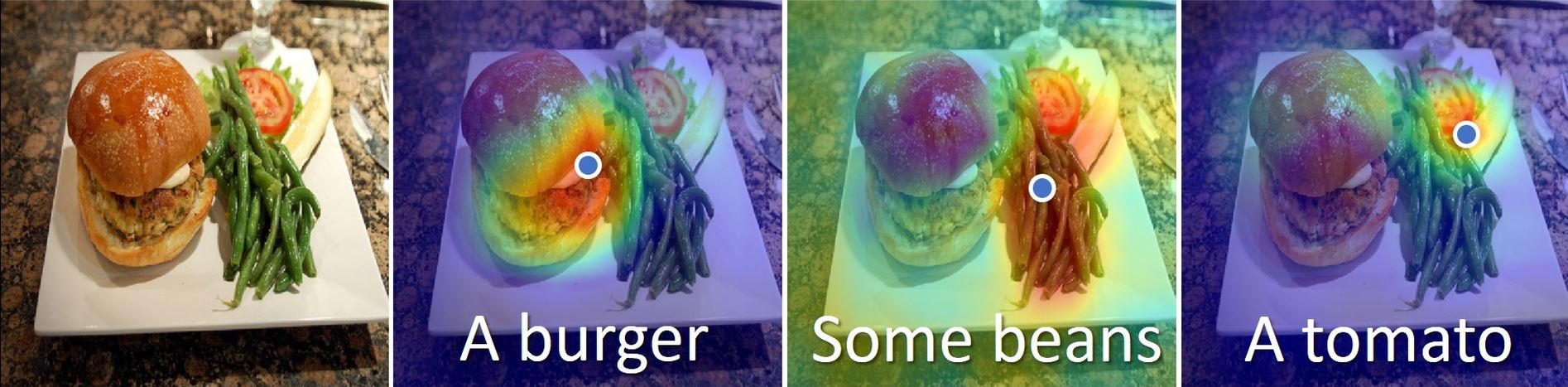}
 \end{center}
 \caption{\textbf{Concept localization with proposed semantic-visual embedding}. Not only does our deep embedding allows cross-modal retrieval with state-of-the-art performance, but it can also associate to an image, \eg, the hamburger plate on the left, a localization heatmap for any text query, as shown with overlays for three text examples. The circled blue dot indicates the highest peak in the heatmap.}
 \label{fig:BigPicture}
\end{figure}

Recent works have demonstrated how such deep representations of images and texts can be jointly leveraged to build \textit{visual-semantic embeddings} \cite{Frome2013,Karpathy2015,Kiros2014,ren2016joint}. The ability to map natural images and texts in a shared representation space where geometry (distances and directions) might be interpreted is a powerful unifying paradigm. Not only does it permit to revisit visual recognition and captioning tasks, but it also opens up new usages, such as cross-modal content search or generation.  

One popular approach to semantic-visual joint embedding is to connect two mono-modal paths with one or multiple fully connected layers \cite{Kiros2014,Karpathy2015,Wang2017,Faghri2017,Ba2016}: A visual path based on a pre-trained convolutional neural network (CNN) and a text path based on a pre-trained recurrent neural network (RNN) operating on a given word embedding. Using aligned text-image data, such as images with multiple captions from \coco~dataset \cite{lin2014microsoft}, final mapping layers can be trained, along with the optional fine-tuning of the two branches. 
Building on this line of research, we investigate new pooling mechanisms in the visual path. Inspired by recent work on weakly supervised object localization \cite{Zhou2016,Durand2016}, we propose in particular to leverage selective spatial pooling with negative evidence proposed in \cite{Durand2016} to improve visual feature extraction without resorting, \eg, to expensive region proposal strategies.
Another important benefit of the proposed joint architecture is that, once trained, it allows localization of arbitrary concepts within arbitrary images: Given an image and the embedding of a text (or any point of the embedding space), we propose a mechanism to compute a localization map, as demonstrated in Fig.\ \ref{fig:BigPicture}.   

The proposed modification to current approaches, along with additional design and training specifics, leads to a new system whose performance is assessed on two very different tasks. We first establish new state-of-the-art performance on cross-modal matching, effectively composed of two symmetric sub-tasks: Retrieving captions from query images and vice-versa. Without additional fine-tuning, our model with its built-in concept localization mechanism also outperforms existing work on the ``pointing game'' sentence-grounding task. With its state-of-the-art performance and its mechanism to localize even unseen concepts, our system opens up new opportunities for multi-modal content search.
The rest of the paper is organized as follows. We discuss in Section \ref{sec:related} the related works, on semantic-visual embedding and on weak supervised localization, and position our work. 
Section \ref{sec:approach} is dedicated to the presentation of our own system, which couples selective spatial pooling with recent architectures and which relies on a triplet ranking loss based on hard negatives. We also show how it can be equipped with a concept localization module by exploiting without pooling the last feature maps in the visual path. More details on the system and its training are reported in Section \ref{sec:expe}, along with various experiments. On the competitive cross-modal retrieval task, our system is shown to outperform current state-of-the-art by a good margin. On the recently proposed task of pointing game, its localization mechanism offers new state-of-the-art performance with no need for retraining. In Section \ref{sec:conclusion}, we finally summarize the achievements of our work and outline perspectives.

\section{Related Work and Paper Positioning }\label{sec:related}
Deep learning offers powerful ways to embed raw data into high dimensional continuous representations that capture semantics. Off-the-shelf pretrained nets are now routinely used to extract versatile deep features from images \cite{krizhevsky2012imagenet,simonyan2014very,He2016} as well as from words and sentences \cite{Mikolov2013,pennington2014glove,chung2014empirical,Lei2017}. There are many strategies either to fine-tune these deep embeddings or to adapt them through new learned projections. In the following, we review learning methods to handle such mono/cross-modal representations, and we also highlight approaches dealing with spatial localization in this context.  

\paragraph{Metric learning for semantic embedding} One way to learn advanced visual representations is to consider the required 
transformation of the raw data 
as a metric learning problem. Several methods have been proposed to learn such metrics. In pairwise approaches, \cite{Xing2002} minimizes the distance within pairs of similar training examples with a constraint on the distance between dissimilar ones. This learning process has been extended to kernel functions as in \cite{mignonpcca}. Other methods consider triplets or quadruplets of images, which are easy to generate from classification training datasets, to express richer relative constraints among groups of similar and dissimilar examples \cite{weinberger09distance,frome2007learning,chechik2010large}.
This kind of learning strategies has been also considered for deep (Siamese) architecture embeddings in the pairwise framework \cite{Cho2005}, and recently extended to triplets \cite{Hof2015}.

To embed words in a continuous space as vector representations, Mikolov \etal's ``word2vec'' is definitively the leading technique \cite{Mikolov2013}. In recent years, several approaches have been developed for learning operators that map sequences of word vectors to sentence vectors including recurrent networks \cite{Hoc1997,chung2014empirical,Lei2017} and convolutional networks \cite{Kim2014}. Using word vector learning as inspiration, \cite{kiros2015skip} proposes an objective function that abstracts the skip-gram word model to the sentence level, by encoding a sentence to predict the sentences around it.

In our work, we adopt most recent and effective deep architectures on both sides, using a deep convolutional network (ResNet) for images \cite{He2016} and a simple recurrent unit (SRU) network \cite{Lei2017} to encode the textual information. Our learning scheme is based on fine-tuning (on the visual side) and triplet-based optimization, in the context of cross-modal alignment that we describe now.  

\paragraph{Learning cross-modal embedding} The Canonical Correlation Analysis (CCA) method is certainly one of the first techniques to align two views of heterogeneous data in a common space \cite{Hot1936}. Linear projections defined on both sides are optimized in order to maximize the cross correlation. Recently, non-linear extensions using kernel (KCCA \cite{kcca2000}) or deep net (DCCA \cite{And2013}) have been proposed. \cite{Wang2016} exploit DCCA strategies for image-text embeddings, while \cite{Yan2015} points out some limitations of this approach in terms of optimization complexity and overfitting and proposes ways to partially correct them. \cite{Eis2017} proposes some CCA-based constraint regularization to jointly train two deep nets passing from one view to the other (text/image).

When considering the specific problem of embedding jointly images and labels (classification context), 
\cite{Wes2011,Frome2013} train models that combine a linear mapping of image features into the joint embedding space with an embedding vector for each possible class label. 
Approaches for the more advanced task of textual image description (captioning) often rely on an encoder/decoder architecture where the encoder consists of a joint embedding \cite{Kiros2014,Karpathy2015}. Other works focus on the sole building of such a joint embedding, to perform image-text  matching and cross-modal retrieval \cite{Frome2013,Faghri2017,Ma2015,salvador2017learning}. 

Our work stems from this latter class. We aim at generating a joint embedding space that offers rich descriptors for both images and texts. We adopt the contrastive triplet loss that follows the margin-based principle to separate the positive pairs from the negative ones with at least a fixed margin. The training strategy with stochastic gradient descent has to be carefully adapted to the cross-modality of the triplets. Following \cite{Faghri2017}, we resort to batch-based hard mining, but we depart from this work, and from other related approaches, in the way we handle localization information.

\paragraph{Cross-modal embedding and localization} Existing works that combine localization and multimodal embedding rely on a two-step process. First, regions are extracted either by a dedicated model, \eg,~ EdgeBox in \cite{Wang2017}, or by a module in the architecture. Then the embedding space is used to measure the similarity between these regions and textual data. \cite{niu2017hierarchical,Karpathy2015} use this approach on the dense captioning task to produce region annotations. It is also used for phrase localization by \cite{Wang2017} where the region with the highest similarity with the phrase is picked.

To address this specific problem of phrase grounding, Xiao \etal~\cite{Xiao2017} recently proposed to learn jointly a similarity score and an attention mask. The model is trained using a structural loss, leveraging the syntactic structure of the textual data to enforce corresponding structure in the attention mask.

In contrast to these works, our approach to spatial localization in semantic-visual embedding is weakly supervised and does not rely on a region extraction model. Instead, we take inspiration from other works on weakly supervised visual localization to design our architecture, with no need for a location-dependent loss.

\paragraph{Weakly supervised localization} The task of generating image descriptors that include localization information has also been explored. A number of weakly supervised object localization approaches extrapolate localization features while training an image classifier, \eg,   \cite{Zhou2016,Durand2016,Dai2016}. The main strategy consists in using a fully convolutional deep architecture that postpones the spatial aggregation (pooling) at the very last layer of the net. It can be used both for classification and for object detection.

We follow the same strategy, but in the context of multi-modal embedding learning, hence with a different goal. In particular, richer semantics is sought (and used for training) in the form of visual description, whether at the scene or at the object level.

\section{Approach}\label{sec:approach}
\begin{figure}[htb]
 \begin{center}
  \centering\includegraphics[width=0.95\linewidth]{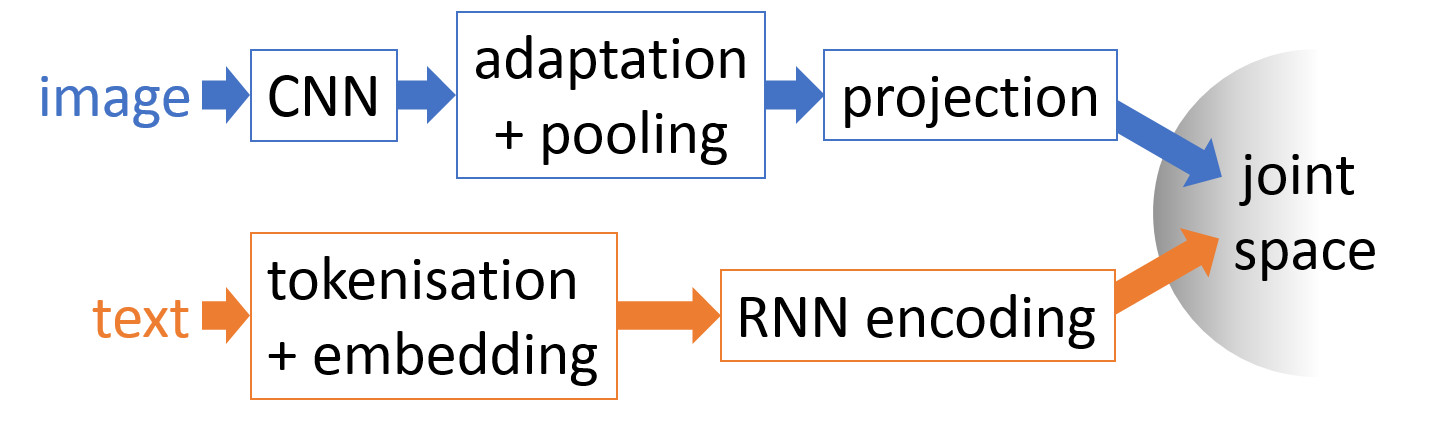}
 \end{center}
 \caption{\textbf{Two-path multi-modal embedding architecture}. Images of arbitrary size and text of arbitrary length pass through dedicated neural networks to be mapped into a shared representation vector space. The visual path (blue) is composed of a fully convolutional neural network (ResNet in experiments), followed by a convolutional adaptation layer, a pooling layer that aggregates previous feature maps into a vector and a final projection to the final output space; The textual path (orange) is composed of a recurrent net running on sequences of text tokens individually embedded with an off-the-shelf map (word2vec in experiments).}

 \label{fig:overall}
\end{figure}

The overall structure of the proposed approach, shown in Fig. \ref{fig:overall}, follows the dual-path encoding architecture of Kiros \etal~\cite{Kiros2014}. We first explain its specifics before turning to its training with a cross-modal triplet ranking loss.    

\subsection{Semantic-visual embedding architecture}

\begin{figure*}[t]
  \begin{center}
  \centering\includegraphics[width=0.80\textwidth]{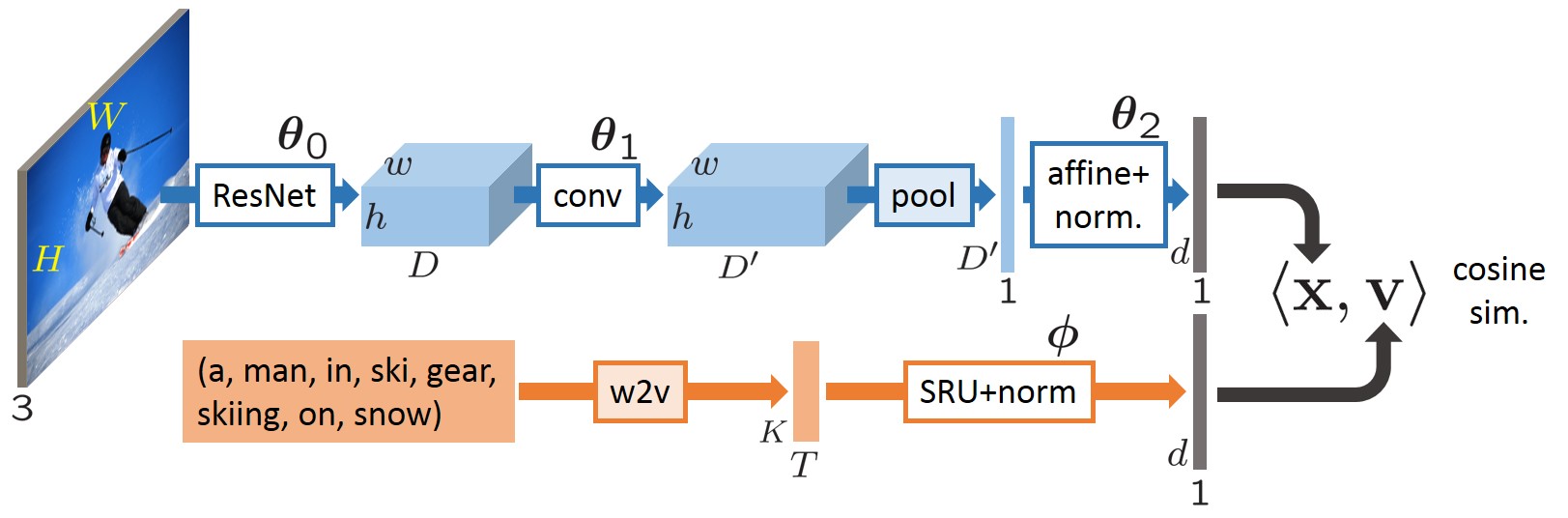}
  \end{center}
  \caption{\textbf{Details of the proposed semantic-visual embedding architecture}. An image of size $3\times W \times H$ is transformed into a unit norm representation $\bx\in\R^{d}$; likewise, a sequence of $T$ tokenized words is mapped to a normalized representation $\bv\in\R^d$. Training will aim to learn parameters $(\mbs \theta_0, \mbs \theta_1,\mbs \theta_2,\mbs \phi)$ such that cross-modal semantic proximity translates into high cosine similarity $\langle\ve x,\ve v\rangle$ in the joint embedded space. Boxes with white background correspond to trainable modules, with parameters indicated on top. In our experiments, the dimensions are $K=620$, $D=2048$ and $D'=d=2400$.}
  \label{fig:full_model}
\end{figure*}

\paragraph{Visual path} In order to accommodate variable size images and to benefit from the performance of very deep architectures, we rely on fully convolutional residual ResNet-152 \cite{He2016} as our base visual network. Its penultimate layer outputs a stack of $D=2048$ feature maps of size $(w,h) = (\frac{W}{32},\frac{H}{32})$, where $(W,H)$ is the spatial size of the input image. These feature maps retain coarse spatial information that lends itself to spatial reasoning in subsequent layers. Following the weakly supervised learning framework proposed by Durand \etal~\cite{Durand2016,durand2017wildcat}, we first transform this stack through a linear adaptation layer of $1\times 1$ convolutions. While in {\sc Weldon} \cite{Durand2016} and in {\sc Wildcat} \cite{durand2017wildcat} the resulting maps are class-related (one map per class in the former, a fixed number of maps per class in the latter), we do not address classification or class detection here. 

Hence we empirically set the number $D'$ of these new maps to a large value, 2400 in our experiments. A pooling {\`a} la {\sc Weldon} is then used, but again in the absence of classes, to turn these maps into vector representations of dimension $D'$. A linear projection with bias, followed by $\ell_2$ normalization accomplishes the last step to the embedding space of dimension $d$.

More formally, the visual embedding path is defined as follows:
\begin{equation}
\ve I \xmapsto{f_{\mbs\theta_0}} 
\ve F \xmapsto{g_{\mbs\theta_1}} 
\ve G \xmapsto{\mathrm{sPool}} 
\ve h \in \R	^{D'} \xmapsto{p_{\mbs \theta_2}}  
\bx \in \R^{d}, 
\label{eq:pipe}
\end{equation}
where: $\ve I \in (0,255)^{W\times H\times 3}$ is the input color image, $f_{\mbs\theta_0}(\ve I)\in\R_+^{w\times h\times D}$ is the output of ResNet's {\tt conv5} parematrized by weights in $\mbs\theta_0$, $g_{\mbs\theta_1}$ is a convolution layer with $|\mbs\theta_1|= D\times D'$ weights and with activation in $\R^{w\times h\times D'}$, $\mathrm{sPool}$ is the selective spatial pooling with negative evidence defined in \cite{Durand2016}:
\begin{equation}
\ve h[k] = \max \ve G[:,:,k] + \min \ve G[:,:,k],~k=1\cdots D',
\label{eq:pool}
\end{equation}
and $p_{\mbs \theta_2}$ is an $\ell_2$-normalized affine function
\begin{equation}
p_{\mbs \theta_2}(\ve h) = \frac{A\ve h + \ve b}{\|A\ve h + \ve b\|_2},
\label{eq:linear}
\end{equation}
where $\mbs\theta_2=(A,\ve b)$ is of size $(D'+1)\times d$. We shall denote $\ve x = F(\ve I;\mbs\theta_{0:2})$ for short this visual embedding. 

\paragraph{Textual path} The inputs to this path are tokenized sentences (captions), \ie, variable length sequences of tokens $S=(s_1 \cdots s_T)$. Each token $s_t$ is turned into a vector representation $\ve s_t\in\R^{K}$ by the pre-trained word2vec embedding \cite{Mikolov2013} of size $K=620$ used in \cite{kiros2015skip}. Several RNNs have been proposed in the literature to turn such variable length sequences of (vectorized) words into meaningful, fixed-sized representations. In the specific context of semantic-visual embedding, \cite{Kiros2014,Faghri2017} use for instance gated recurrent unit (GRU) \cite{chung2014empirical} networks as text encoders. Based on experimental comparisons, we chose to encode sentences with the simple recurrent unit (SRU) architecture recently proposed in \cite{Lei2017}. Since we train this network from scratch, we take its output, up to $\ell_2$ normalization, as the final embedding of the input sentence. There is no need here for an additional trainable projection layer.  

Formally, the textual path reads:
\begin{equation}
S \xmapsto{\mathrm{w2v}} \ve S \xmapsto{\mathrm{normSRU}_{\mbs\phi}} \ve v \in \R^{d} ,
\end{equation}
where $\ve S = \mathrm{w2v}(S) = \R^{K\times T}$ is an input sequence of text tokens vectorized with word2vec and $\ve v$ is the final sentence embedding in the joint semantic-visual space, obtained after $\ell_2$-normalizing the output of SRU with parameters $\mbs\phi$. 

\subsection{Training} The full architecture is summarized in Fig. \ref{fig:full_model}. The aim of training it is to learn the parameters $\mbs\theta_{0:2}$ of the visual path, as well as all parameters $\mbs\phi$ of the SRU text encoder. The goal is to create a joint embedding space for images and sentences such that closeness in this space can be interpreted as semantic similarity. This requires cross-modal supervision such that image-to-text semantic similarities are indeed enforced.\footnote{Note that mono-modal supervision can also be useful and relatively easier to get in the form, \eg,~of categorized images or of categorized sentences. Both are indeed used implicitly when relying on pre-trained CNNs and pre-trained text encoders. It is our case as well as far as the visual path is concerned. However, since our text encoder is trained from scratch, the only pure text (self-)supervision we implicitly use lies in the pre-training of word2vec.}
  
\paragraph{Contrastive triplet ranking loss} Following \cite{Kiros2014}, we resort to a contrastive triplet ranking loss. Given a training set $\mathcal{T} = \big\{(\ve I_n,S_n)\big\}_{n=1}^N$ of aligned image-sentence pairs -- the sentence describes (part of) the visual scene -- the empirical loss to be minimized takes the form:
\begin{align}
\mathcal{L}(\mbs\Theta;\mathcal{T}) =  \frac{1}{N}\sum_{n=1}^N \Big( & \sum_{m\in C_n}  \mathrm{loss}(\ve x_n,\ve v_n,\ve v_m) \nonumber \\
 + & \sum_{m\in  D_n}  \mathrm{loss}(\ve v_n,\ve x_n,\ve x_m)\Big),
\label{eq:loss}
\end{align}   
where $\mbs\Theta =(\mbs\theta_0,\mbs\theta_1,\mbs\theta_2,\mbs\phi)$ are the parameters to learn, $\ve x_n = F(\ve I_n;\mbs\theta_{0:2})$ is the embedding of image $n$, $\ve v_n = \mathrm{normSRU}_{\mbs\phi}(\mathrm{w2v}(S_n))$ is the embedding of sentence $n$, $\{S_m\}_{m\in C_n}$ is a set of sentences unrelated to $n$-th image, $\{\ve I_m\}_{m\in D_n}$ is a set of images unrelated to $n$-th sentence. The two latter sets are composed of negative (``constrastive'') examples. The triplet loss is defined as:
\begin{equation}
\mathrm{loss}(\ve y,\ve z,\ve z') = \max\big\{0,\alpha - \si{\ve y }{\ve z} + \si{\ve y }{\ve z'}\big\}, 
\label{eq:triplet}
\end{equation} 
with $\alpha>0$ a margin. It derives from triplet ranking losses used to learn metrics and to train retrieval/ranking systems. The first argument is a ``query'', while the second and third ones stand respectively for a relevant (positive) answer and an irrelevant (negative) one. The loss is used here in a similar way, but with a multimodal triplet. In the first sum of Eq. \ref{eq:loss}, this loss encourages the similarity, in the embedding space, of an image with a related sentence to be larger by a margin to its similarity with irrelevant sentences. The second sum is analogous, but centered on sentences.   

\paragraph{Mining hard negatives} In \cite{Kiros2014,Karpathy2015}, contrastive examples are sampled at random among all images (resp. sentences) in the mini-batch that are unrelated to the query sentence (resp. image). Faghri \etal~\cite{Faghri2017} propose instead to focus only on the hardest negatives. We follow the same strategy:
For each positive pair in the batch, a single contrastive example is selected in this batch as the one that has the highest similarity with the query image/sentence while not being associated with it. This amounts to considering the following loss for the current batch $\mathcal{B}=\big\{(\ve I_n,S_n)\big\}_{n\in B}$:
\begin{align}
\mathcal{L}(\mbs\Theta;\mathcal{B}) =  
\frac{1}{|B|}\sum_{n\in B} \Big( & \max_{m\in C_n\cap B}  \mathrm{loss}(\ve x_n,\ve v_n,\ve v_m) \nonumber \\
+ & \max_{m\in D_n\cap B}  \mathrm{loss}(\ve v_n,\ve x_n,\ve x_m)\Big).
\label{eq:loss2}
\end{align} 
Beyond its practical interest, this mining strategy limits the amount of gradient averaging, making the training more discerning.

\subsection{Localization from embedding}

As described in Section \ref{sec:related}, several works on weak supervised localization \cite{Zhou2016,Durand2016} combine fully convolutional architectures with specific pooling mechanisms such that the unknown object positions in the training images can be hypothesized. 
This localization ability derives from the activation maps of the last convolutional layer. Suitable linear combinations of these maps can indeed provide one heatmap per class. 
   
Based on the pooling architecture of \cite{Durand2016} which is included in our system and without relying on additional training procedures, we derive the localization mechanism for our semantic-visual embedding. Let's remind that in our case, the number of feature maps is arbitrary since we are not training on a classification task but on a cross-modal matching one. Yet, one can imagine several ways to leverage these maps to try and map an arbitrary vector of the joint embedding space into an arbitrary input image. When this vector is the actual embedding of a word or sentence, this spatial mapping should allow localizing the associated concept(s) in the image, if present. Ideally, a well-trained joint embedding should allow such localization even for concepts that are absent from the training captions.

\begin{figure}[t]
  \centering
  \includegraphics[width=\linewidth]{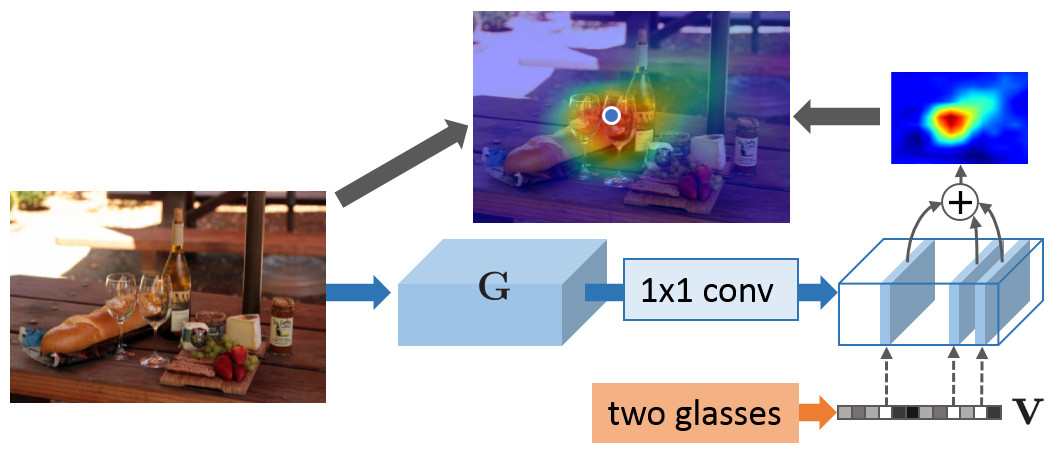}
  \caption{\textbf{From text embedding to visual localization}. Given the feature maps $\ve G$ associated to an image by our semantic-visual architecture and the embedding of a sentence, a heatmap can be constructed: Learned projection matrix $A$ serves as a $1\times 1$ convolution; Among the $d$ maps thus generated, the $k$ ones associated with the largest among the $d$ entries of $\ve v$ are linearly combined. If the sentence relates to a part of the visual scene, like ``two glasses''  in this example, the constructed heatmap should highlight the corresponding location. Blue dot indicates the heat maximum.}
  \label{fig:localize}
\end{figure}

To this end, we propose the following localization process (Fig. \ref{fig:localize}). Let $\ve I$ be an image and $\ve G$ its associated $D'$ feature maps (Eq. \ref{eq:pipe}). This stack is turned into a stack $\ve G'\in\R^{w\times h\times d}$ of $d$ heatmaps using the linear part of the projection layer $p_{\mbs \theta_2}$:\footnote{In other words, the pooling is removed. Bias and normalization being of no incidence on the location of the peaks, they are ignored.} 
\begin{equation}
\ve G'[i,j,:]= A \ve G[i,j,:],~\forall (i,j)\in \llbracket 1,w\rrbracket \times \llbracket 1, h\rrbracket,
\end{equation}
which is a $1\times 1$ convolution. 
Given $\ve v\in\R^d$ the embedding of a word or sentence (or any unit vector in the embedded space) and $K(\ve v)$ the set of the indices of its $k$ largest entries, the 2D heatmap $\ve H \in \R^{w\times h}$ associated with the embedded text $\ve v$ in image $\ve I$ is defined as:
\begin{equation}
\ve H = \sum\nolimits_{u\in K(\ve v)} \big| \ve v[u] \big|\times \ve G'[:,:,u].
\end{equation} 
In the next section, such heatmaps will be shown in false colors, overlaid on the input image after suitable resizing, as illustrated in Figs.\ \ref{fig:BigPicture} and \ref{fig:localize}. Note that \cite{selvaraju2017} also proposes to build semantic heatmaps as weighted combinations of feature maps, but with weights obtained by back-propagating the loss in their task-specific network (classification or captionning net). Such heatmaps help visualize which image regions explain the decision of the network for this task.
  
\section{Experiments}\label{sec:expe}
Starting from images annotated with text, we aim at producing rich descriptors for both image and text that live in the same embedding space.
Our model is trained on the \coco~  dataset, and benchmarked on two tasks. To evaluate the overall quality of the model we use cross-modal retrieval, and to assess its localization ability we tackle visual grounding of phrases.

\begin{table*}[ht]
\begin{center}
  \rowcolors{2}{gray!20}{white}
\begin{tabular}{r c c c c c c  c c c c c } \rowcolor{gray!40}
 & visual & & \multicolumn{4}{c}{caption retrieval} & & \multicolumn{4}{c}{image retrieval} \\ \rowcolor{gray!40}
model &  backend && R@1 & R@5 & R@10 & Med. r && R@1 & R@5 & R@10 & Med. r \\
Embedding network \cite{Wang2017} & VGG && 50.4 & 79.3 & 89.4 & - && 39.8 & 75.3 & 86.6 & - \\
2-Way Net \cite{Eisenschtat2016} & VGG && 55.8 & 75.2 & - & - && 39.7 & 63.3 & - & - \\
LayerNorm \cite{Ba2016} & VGG && 48.5 & 80.6 & 89.8 & 5.1 && 38.9 & 74.3 & 86.3 & 7.6 \\
VSE++  \cite{Faghri2017} & R152 && 64.6 & - & 95.7 & 1  && 52.0 & - & 92.0 & 1  \\
Ours & R152 && \textbf{69.8} & \textbf{91.9} & \textbf{96.6} & 1  && \textbf{55.9} & \textbf{86.9} & \textbf{94.0} & 1
\end{tabular}
\end{center}
  \caption{\textbf{Cross-modal retrieval results on \coco.} On both caption retrieval from images and image retrieval from captions, the proposed architecture outperforms the state-of-the-art systems. It yields an R@1 relative gain of 38\% (resp. 40\%) with respect to best published results \cite{Wang2017} on cross-modal caption retrieval (resp. image retrieval), and 8\% (resp 7.5\%) with respect to best online results \cite{Faghri2017}.}  
 \label{tab:retrieval}
\end{table*}

\subsection{Training}

\paragraph{Datasets} To train our model, we used the \coco~ dataset \cite{lin2014microsoft}\footnote{\url{http://cocodataset.org}}. This dataset contains 123,287 images (train+val), each of them annotated with 5 captions. It is originally split into a training set of 82,783 images and a validation set of 40,504 images. The authors of \cite{Karpathy2015} proposed another split (called rVal in the rest of the paper) keeping from the original validation set 5,000 images for validation and 5,000 for testing and using the remaining 30,504 as additional training data. To make our results comparable, we trained a model using each split. For evaluation, we also use the \coco~  dataset, complemented with the annotations from \genome~dataset \cite{krishna2016visual}\footnote{\url{http://visualgenome.org/}} to get localization ground-truth when needed. 

\paragraph{Image pipeline} The image pipeline is pre-trained on its own in two stages. We start from original ResNet-152 \cite{He2016} pre-trained on ImageNet classification task. Then, to initialize the convolutional adaptation layer $g_{\mbs\theta_1}$, we consider temporarily that the post-pooling projection is of size $1000$ such that we can train both on ImageNet as well. Once this pre-training is complete, the actual projection layer $p_{\mbs\theta_2}$ onto the joint space is put in place with random initialization, and combined with a 0.5-probability dropout layer. As done in \cite{Faghri2017}, random rectangular crops are taken from training images and resized to a fixed-size square (of size $256\times 256$).   

\paragraph{Text pipeline} To represent individual word tokens as vectors, we used pre-trained word2vec with no further fine-tuning. The SRU text encoder \cite{Lei2017} is trained from scratch jointly with the image pipeline. It has four stacked hidden layers of dimension 2400. Following \cite{Lei2017}, 0.25-probability dropout is applied on the linear transformation from input to hidden state and between the layers. 

\paragraph{Full model training} Both pipelines are trained together with pairs of images and captions, using Adam optimizer \cite{Kingma2014}. Not every part of the model is updated from the beginning. For the first 8 epochs only the SRU (parameters $\mbs\phi$) and the last linear layer of the image pipeline ($\mbs\theta_2$) are updated. After that, the rest of the image pipeline ($\mbs\theta_{0:1}$) is also fine-tuned. The training starts with a learning rate of 0.001 which is then divided by two at every epoch until the seventh and kept fixed after that. Regarding mini-batches, we found in contrast to \cite{Faghri2017} that their size has an important impact on the performance of our system. After parameter searching, we set this size to 160. Smaller batches result in weaker performance while too large ones prevent the model from converging.

\subsection{Comparison to state-of-the-art}

\paragraph{MS-COCO retrieval task} Our model is quantitatively evaluated on a cross-modal retrieval task. Given a query image (resp. a caption), the aim is to retrieve the corresponding captions (resp. image). Since \coco~contains 5 captions per image, recall at $r$ (``R@$r$'') for caption retrieval is computed based on whether at least one of the correct captions is among the first $r$ retrieved ones. The task is performed 5 times on 1000-image subsets of the test set and the results are averaged. 

All the results are reported on Tab.~\ref{tab:retrieval}. We compare our model with recent leading methods. 
As far as we know, the best published results on this task
are obtained by the Embedding Network \cite{Wang2017}. For caption retrieval, we surpass it by (19.4\%,12.6\%,7.2\%) on (R@1,R@5,R@10) in absolute, and by (16.1\%,11.6\%,7.4\%) for image retrieval. Three other methods are also available online, 2-Way Net \cite{Eisenschtat2016}, LayerNorm \cite{Ba2016} and VSE++ \cite{Faghri2017}. The first two are on the par with Embedding Network while VSE++ reports much stronger performance. We consistently outperform the latter, especially in terms of R@1. 
The most significant improvement comes from the use of hard negatives in the loss, without them recall scores are significantly lower (R@1 - caption retrieval: -20,3\%, image retrieval: -16.3\%).

Note that in \cite{Faghri2017}, the test images are scaled such that the smaller dimension is 256 and centrally cropped to $224\times 224$. Our best results are obtained with a different strategy: Images are resized to $400\times 400$ irrespective of their size and aspect ratio, which our fully convolutional visual pipeline allows. When using the scale-and-crop protocol instead, the recalls of our system are reduced by approximately 1.4\% in average on the two tasks, remaining above VSE++ but less so. For completeness we tried our strategy with VSE++, but it proved counterproductive in this case.

\begin{table*}[h]
  \begin{center}
  \rowcolors{2}{gray!20}{white}
\begin{tabular}{r c c c c c  c c c c c } \rowcolor{gray!40}
 & & \multicolumn{4}{c}{caption retrieval} & & \multicolumn{4}{c}{image retrieval} \\ \rowcolor{gray!40}
model && R@1 & R@5 & R@10 & Med. r  && R@1 & R@5 & R@10 & Med. r  \\
Emb. network \cite{Wang2017} && 40.7 & 69.7 & 79.2 & - && 29.2 & 59.6 & 71.7 & - \\
2-Way Net \cite{Eisenschtat2016} && 49.8 & 67.5 & - & - && 36.0 & 55.6 & - & - \\
VSE++  \cite{Faghri2017} && 52.9 & - & 87.2 & 1 && \textbf{39.6} & - & 79.5 & 2 \\
DAN  \cite{nam2016dual} && \textbf{55.0} & \textbf{81.8} & \textbf{89.0} & 1 && 39.4 & 69.2 & 79.1 & 2 \\
Ours (\coco~only) && 46.5 & 72.0 & 82.2 & 2 && 34.9 & 62.4 & 73.5 & 3 
\end{tabular}
\end{center}
  \caption{\patrev{\textbf{Direct transfer to \flickr, with comparison to SoA}. Although cross-validated and trained on \coco~only, our system delivers good cross-modal retrieval performance on \flickr, compared to recent approaches trained on \flickr: It is under the two best performing approaches, but above the two others on most performance measures.}}  
  \label{tab:retrievalflickr}
\end{table*}

\paragraph{Visual grounding of phrases} We evaluate quantitatively our localization module with the pointing game defined by \cite{Xiao2017}. This task relies on images that are present both in \coco~{\tt val} 2014 dataset and in \genome~dataset. The data contains 17,471 images with 86,5582 text region annotations (a bounding box associated with a caption). The task consists in ``pointing'' the region annotation in the associated image. If the returned location lies inside the ground-truth bounding box, it is considered as a correct detection, a negative one otherwise. Since our system produces a localization map, the location of its maximum is used as output for the evaluation.
For this evaluation, the number of feature maps from $\ve G'$ that are used to produce the localization map was set through cross-validation to $k=180$ (out of 2400). We keep this parameter fixed for all presented visualizations. 

\begin{figure}[tb]
\centering
\includegraphics[width=\columnwidth]{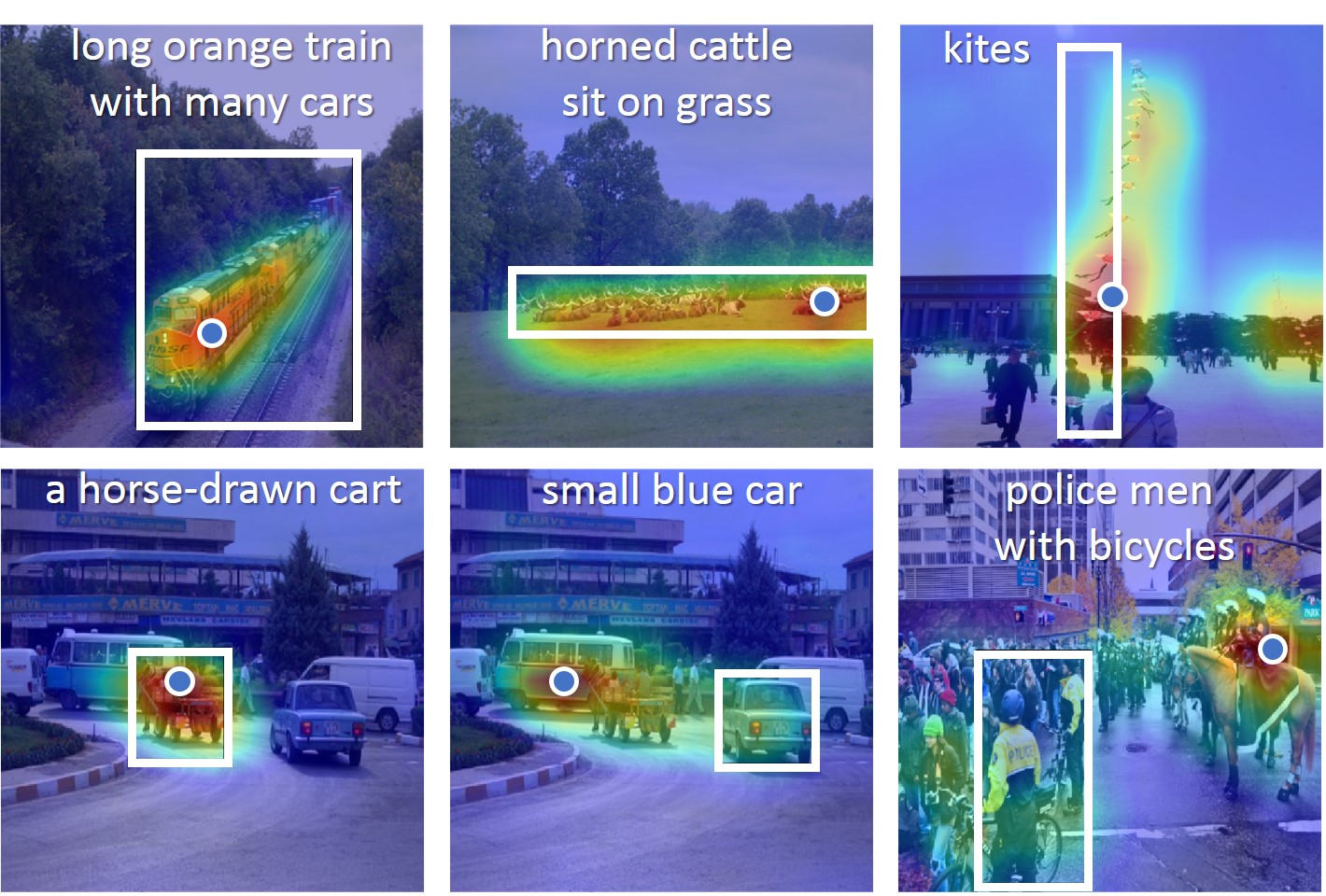}
\caption{\textbf{Pointing game examples}. Images from the \genome~dataset overlaid with the heatmap localizing the input text according to our system. The white box is the ground-truth localization of the text and the blue dot marks the location predicted by our model for this text. The first four predictions are correct, unlike the last two ones. In the last ones, the heatmap is nonetheless active inside the ground-truth box.}
\label{fig:point_loc}
\end{figure}

The quantitative results are reported in Tab.~\ref{tab:point} and some visual examples are shown in Fig.~\ref{fig:point_loc}. We add to the comparison a baseline that always outputs the center of the image as localization, leading to a surprisingly high accuracy of 19.5\%.  
Our model, with an accuracy of 33.8\%, offers absolute (resp. relative) gains of 9.4\% (resp. 38\%) over \cite{Xiao2017} and of 14\% (resp. 73\%) over the trivial baseline.  

\paragraph{MS-COCO localization and segmentation} 
Following the evaluation scheme for \cite{Xiao2017}, we obtain similar semantic segmentation performance (namely mAP scores of 0.34, 0.24 and 0.15 for
IoU@0.3, IoU@0.4 and IoU@0.5 resp.), while our localization module does not benefit from a training to structure the heatmaps.  
We also performed pointwise object localization on \coco~using the bounding box annotation, obtaining 57.4 mAP, an improvement of 4\% compared to \cite{
durand2017wildcat}.

\begin{table}[t]
\begin{center}
\rowcolors{2}{gray!20}{white}
\begin{tabular}{lcc}\rowcolor{gray!40}
Model && Accuracy \\
``center'' baseline && 19.5 \\
Linguistic structure \cite{Xiao2017} && 24.4 \\
Ours ({\tt train 2017}) && 33.5 \\
Ours ({\tt rVal}) && 33.8 \\
\end{tabular}
\end{center}
\caption{\textbf{Pointing game results}. Our architecture outperforms the state-of-the-art system \cite{Xiao2017} by more than 9\% in accuracy, when trained with either {\tt train} or {\tt rVal} split from \coco.}
\label{tab:point}
\end{table}

\subsection{Further analysis}

\paragraph{Transfer to Flickr30K} We propose to investigate how our model trained on \coco~may be transferred as such to other datasets, namely \flickr~here. We report the results in Tab. \ref{tab:retrievalflickr}. Not surprisingly, our performance is below the best systems \cite{nam2016dual,Faghri2017} trained on \flickr. Yet, while not being trained at all on \flickr, it outperforms on almost all measures two other recent approaches trained on \flickr~\cite{Eisenschtat2016,Wang2017}. Note that \textit{fine-tuning} our system on \flickr~makes it outperform all, including \cite{nam2016dual,Faghri2017}, by a large margin (not reported in Table for the sake of fairness).\footnote{We chose to keep the architecture used on \coco\ as it is and to experiment with transfer and fine-tuning. An actual evaluation on \flickr\ would require 
cross-validation of the various hyper-parameters. This dataset being substantially smaller than \coco, such a task is challenging given the size of our architecture with its 2400 new feature maps and its large final embedding dimension of 2400.}

\begin{figure*}[t]
\centering
\includegraphics[width=0.68\textwidth]{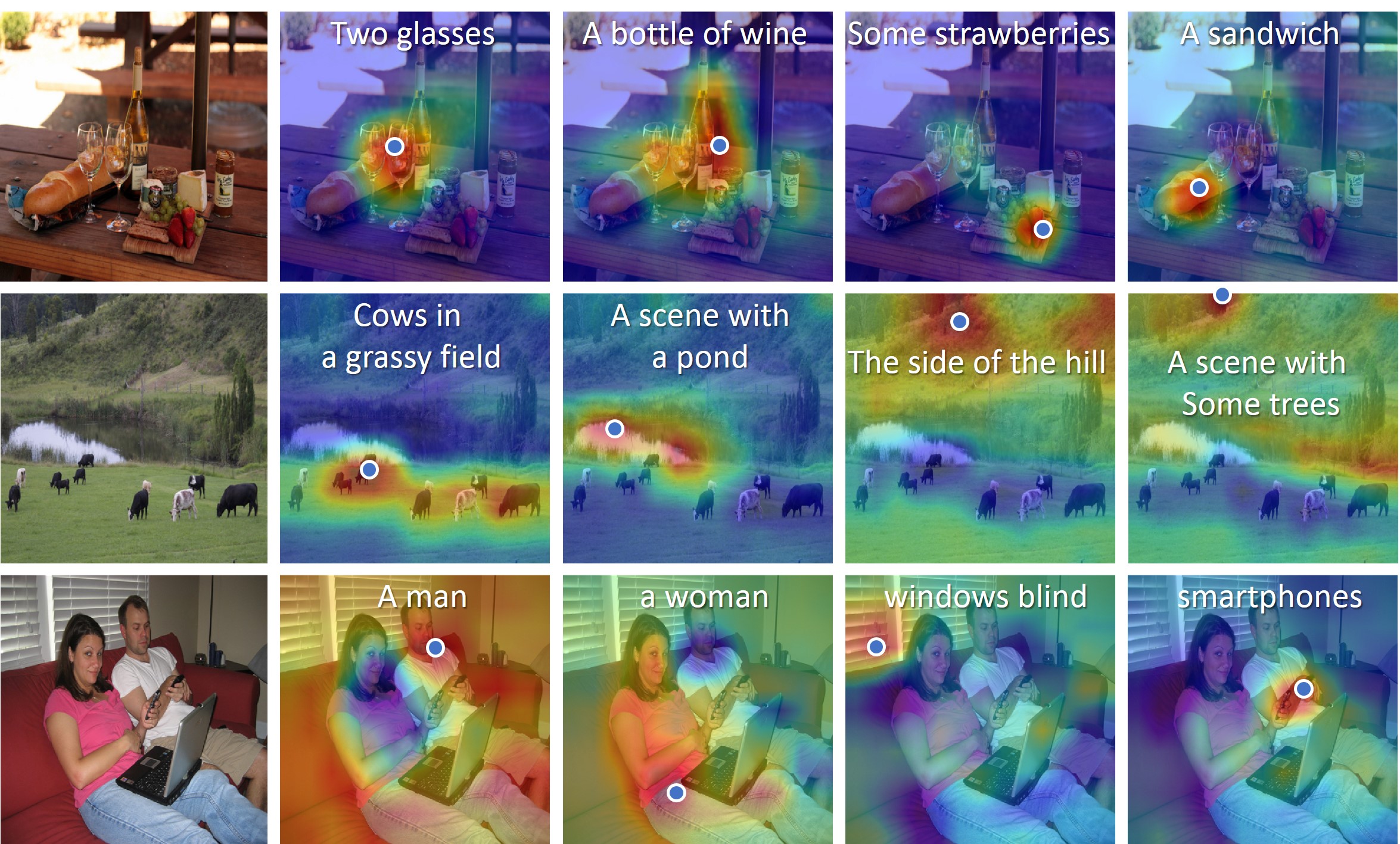}
\caption{\textbf{Localization examples}. The first column contains the original image, the next columns show as overlays the heatmaps provided by the localization module of our system for different captions (superimposed). In each image the circled blue dot marks the maximum value of the heatmap.}
\label{fig:local_notseen}
\end{figure*}

\begin{figure}[htb]
\centering
\includegraphics[width=\columnwidth]{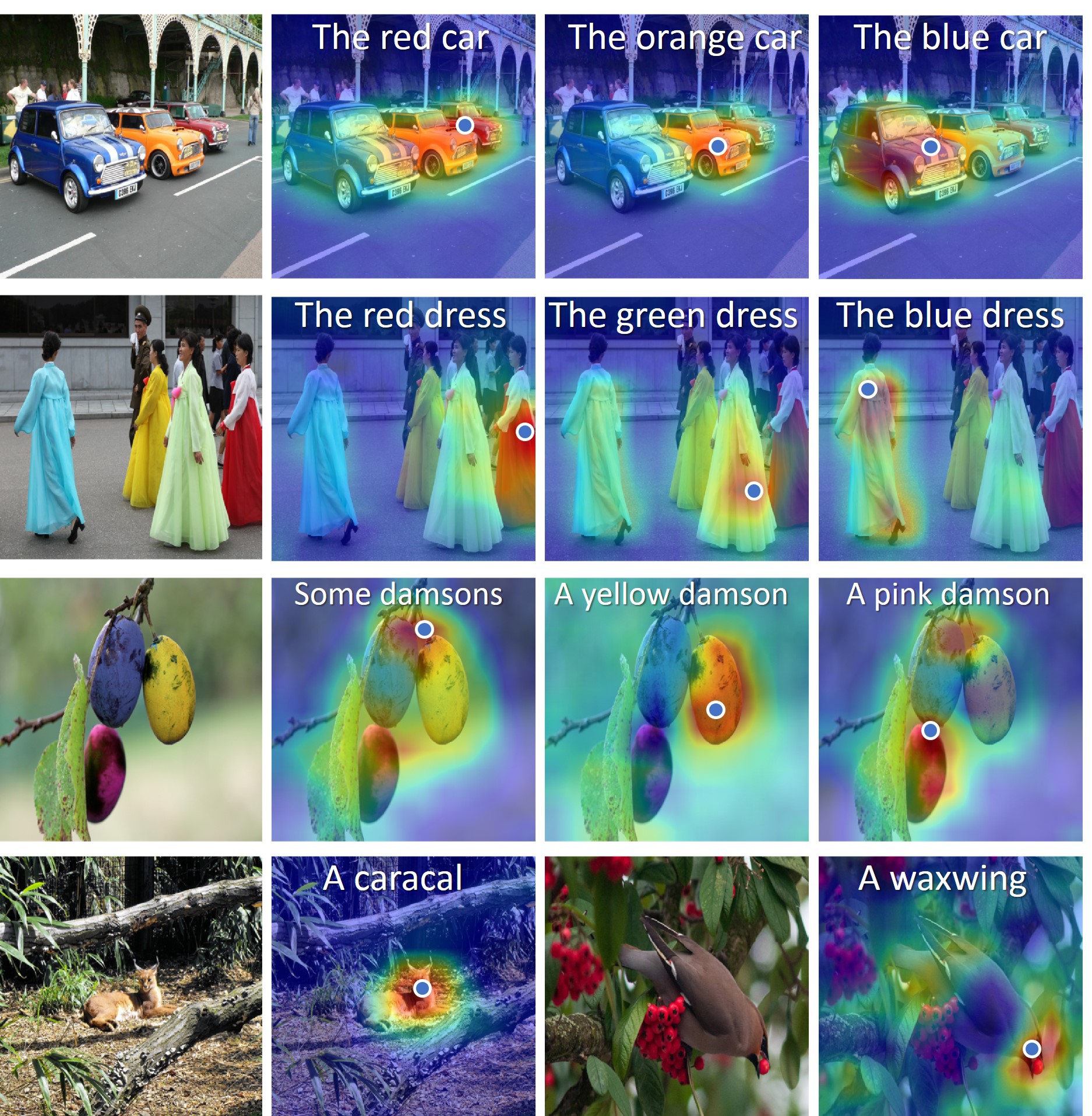}
\caption{\textbf{Toward zero-shot localization}. The first three rows show the ability to differentiate items according to their colors, even if, as in third example, the colors are unnatural and the concept has not been seen at training. This example, and the two last ones could qualify as ``zero-shot localization'' as {\tt damson}, {\tt caracal}, and {\tt waxwing} are not present in \coco~train set.}
\label{fig:zero}
\end{figure}

\paragraph{Towards zero-shot localization} The good performance we obtain in the pointing game highlights the ability of our system to localize visual concepts based on their embedding in the learned joint space. We illustrate further this strength of the system with additional examples, like the one already presented in Fig.\ \ref{fig:BigPicture}. We show in Fig.\ \ref{fig:local_notseen} the heatmaps and associated localizations for home-brewed text ``queries" on images from \coco~test set. Going one step further, we conducted similar experiments with images from the web and concepts that were checked \textit{not to appear in any of the training captions}, see Fig.\ \ref{fig:zero}.

\paragraph{Changing pooling}\label{par:ablation}
One of the key elements of the proposed architecture is the final pooling layer, adapted from {\sc Weldon} \cite{Durand2016}. To see how much this choice contributes to the performance of the model, we tried instead the Global Average Pooling (GAP) \cite{Zhou2016} approach. 
With this single modification, the model is trained following the exact same procedure as the original one. 
This results in less good results: For caption retrieval (resp. image retrieval), it incurs a loss of 5.3\% for R@1 (resp. 4.7\%) for instance,
and a loss of 1.1\% in accuracy in the pointing game.

\section{Conclusion}\label{sec:conclusion}
We have presented a novel semantic-visual embedding pipeline that leverages recent architectures to produce rich, comparable descriptors for both images and texts. The use of a selective spatial pooling at the very end of the fully convolutional visual pipeline allows us to equip our system with a powerful mechanism to locate in images the regions corresponding to any text. Extensive experiments show that our model achieves high performance on cross-modal retrieval tasks as well as on phrases localization. We also showed first qualitative results of zero-shot learning, a direction towards which our system could be pushed in the future with, among others, a deeper exploitation of language structure and of its visual grounding.    

\paragraph{Acknowledgments} This research was supported by the ANR-16-CE23-0006 and IBM Montpellier Cognitive Systems Lab with the loan of a PowerAI server. 

\clearpage
{\small
\bibliographystyle{ieee}
\bibliography{egbib}
}

\end{document}